Are Multimodal LLMs Ready for Clinical Dermatology? A Real-World Evaluation in Dermatology

Roy Jiang[1,2], Hyunjae Kim[2], Zhenyue Qin[2], Morten Lee[3], Margaret MacGibeny[1], Ailish Hanly[1], Angela Sadlowski[3], Shanin Chowdhury[3], Xuguang Ai[2], Jeffrey Gehlhausen[1,#], Qingyu Chen[2,*,#]

1. Department of Dermatology, Yale School of Medicine, Yale University, New Haven, CT, USA

2. Department of Bioinformatics and Data Science, Yale School of Medicine, Yale University, New Haven, CT, USA

3. Yale School of Medicine, Yale University, New Haven, CT, USA

[#]Co-senior author

Corresponding author:

*Qingyu Chen, qingyu.chen@yale.edu

Figure count: 3

Table count: 2

## ABSTRACT

**Background:** Multimodal large language models (MLLMs) have demonstrated promise on publicly available dermatology benchmarks. However, benchmark performance may not generalize to real-world dermatologic decision-making. Most prior evaluations rely on curated image-only tasks, often using dermoscopic or single-lesion images obtained under standardized conditions. In contrast, real-world dermatology consultations commonly involve photographs captured in primary care or emergency department settings, rarely with dermoscopy. Consultations are accompanied by brief requests with context written by referring clinicians that may sometimes be misleading. Cases are reviewed by dermatologists, who triage patients for possible urgent evaluation based on likely differential diagnoses.

**Methods:** To quantify this benchmark-to-bedside gap, we evaluated four open-weight MLLMs (InternVL-Chat v1.5, LLaVA-Med v1.5, SkinGPT4 and MedGemma-4B-Instruct) and one commercial MLLM (GPT-4.1) across three publicly available dermatology datasets and a retrospective multi-site hospital-based dermatology consultation cohort comprising 5,811 cases and 46,405 clinical images. The real-world cohort paired clinical photographs with referring clinician consultation notes and dermatologist-authored differential diagnoses. Models were evaluated on two clinically relevant tasks: free-text differential diagnosis generation and severity-based triage of urgent dermatologic conditions. Diagnostic performance was assessed using BERTScore, LLM-based adjudication, and independent review by board-certified or board-eligible dermatologists for clinical intent, factuality, and specificity.

**Results:** Diagnostic performance was modest on public datasets and declined substantially in the real-world consultation cohort. On public benchmarks, top-3 diagnostic accuracy reached 26.55% for the best open-weight model and 42.25% for GPT-4.1. On real-world consultation cases using images alone, top-3 diagnostic accuracy fell to 1.50%-13.35% among open-weight models and 24.65% for GPT-4.1. Incorporating clinical context from referring clinician requests improved performance across all models, increasing top-3 diagnostic accuracy up to 28.75% among open-weight models and 38.93% for GPT-4.1. However, model outputs were highly sensitive to incomplete or erroneous consultation context, leading to substantial diagnostic variability across evaluation settings. For severity-based triage of urgent dermatologic conditions, models achieved moderate sensitivity (above 60%), suggesting potential utility for screening but insufficient reliability for clinical deployment overall.

**Conclusions:** These findings demonstrate that benchmark performance substantially overestimates the real-world clinical capability of current dermatology MLLMs. More importantly, an important barrier to safe deployment appears to not be downstream reasoning, but rather the fragile integration of clinical context as well as the insufficient grounding of diagnosis-relevant visual features under realistic consultation conditions.

## INTRODUCTION

Medical multimodal artificial intelligence has emerged as a promising tool to assist clinical decision-making, which often depends on integrating heterogeneous data sources including medical images, clinical notes, and structured health records.[1,2] Building on advances in large language models (LLMs),[3–5] multimodal large language models (MLLMs) have recently emerged that can jointly process visual and textual inputs while generating natural-language outputs.[6–10] Unlike earlier supervised approaches for narrow classification tasks, Recent MLLMs have shown promising zero-shot performance, without task-specific training, across a range of biomedical and clinical benchmarks: medical image comprehension, disease diagnosis, and question answering. In addition, MLLMs show generative and reasoning capabilities that could potentially support complex clinical workflows, such as clinical assessment and management planning.[9,10]

Dermatology represents a particularly promising domain for MLLMs; dermatologic diagnosis relies heavily on interpreting visual morphological features in conjunction with contextual clinical information. Clinical dermatology images do not require specialized equipment to capture and are abundant within the EHR. Accordingly, artificial intelligence (AI) approaches in dermatology have evolved rapidly to reduce manual burden in clinical workflows such as assisting diagnosis and triage. Early systems relied on supervised convolutional neural network–based classifiers that performed direct diagnosis from dermatologic images.[11] Subsequent approaches shifted toward large-scale representation learning; for example, PanDerm trained CLIP-style vision encoders on large dermatology image datasets to learn generalizable visual representations.[12,13] More recently, dermatology-focused MLLMs have emerged, including systems such as SkinGPT-4, which combine dermatology-trained vision encoders with large language model backbones.[14] Additional dermatology-specific multimodal models have also been proposed using public or synthetic dermatology datasets.[15–17] Alongside these systems, general-domain and medically adapted MLLMs, including early pioneering models such as LLaVA-Med and more recent models such as MedGemma, have been widely used to develop and evaluate for dermatologic applications.[18,19] Collectively, these studies demonstrate promising performance of MLLMs on standard dermatology benchmarks.[5,20,21]

Despite rapid progress, existing benchmark evaluations do not reflect the realities of clinical dermatology workflow.[15–17,22–26] In routine consultations, referring clinicians—often without specialized dermatology training—submit clinical photographs to dermatologists together with contextual information they consider relevant, including patient demographics, symptoms, medical history, available laboratory or pathology results, and their own diagnostic impression. Two characteristics of this process are particularly important. First, consultation photographs are typically captured using cameras or mobile devices rather than with dermoscopy and may vary substantially in lighting, framing, and overall image quality. Second, consultation notes from referring clinicians may be incomplete or inaccurate, reflecting variability in dermatologic expertise. Dermatologists therefore synthesize heterogeneous inputs (clinical photographs of variable quality alongside imperfect clinical context) to generate an open-ended differential diagnosis and determine appropriate triage, that is, whether a case requires in-person dermatologic evaluation.

Critically, current evaluations of MLLMs in dermatology do not capture realistic scenarios requiring photography-based clinical review. Most models are evaluated on public datasets composed of curated, single-lesion images obtained under standardized conditions, such as dermoscopy images, which differ substantially from the variability, distractors, and lighting conditions present in real clinical images.[15–17,22–26] These datasets also support image-based classification tasks only and lack paired clinical narratives or consultation-style notes. Consequently, evaluation tasks primarily emphasize category-based diagnosis or multiple-choice question answering, which differs fundamentally from open-ended differential diagnosis generation required in practice.[27,28] Furthermore, the distinction between dermoscopy images and clinical photography is rarely considered; many studies aggregate these modalities within the same evaluation framework, obscuring modality-specific performance differences. As a result, whether performance on existing dermatology benchmarks translates to real-world clinical consultations remains unclear.

In this study, we evaluate open-weight MLLMs and one commercial comparator under conditions designed to mirror real-world dermatology consultation workflows (Figure 1, Figure S1). Specifically, we assess five MLLMs—InternVL-Chat v1.5 (2B parameters), LLaVA-Med v1.5 (Mistral-7B backbone), SkinGPT-4 (LLaMA v2 13B backbone), MedGemma-4B (instruction-tuned) and GPT-4.1 (commercial model)—across both publicly available dermatology datasets and a large multi-site hospital dermatology consultation cohort comprising 5,811 cases and 46,405 clinical images.[18,19,29,30] Models were evaluated on two clinically relevant tasks: open-ended differential diagnosis generation and triage of high-severity dermatologic conditions. Performance was assessed using automated metrics together with manual review by three board-certified/eligible dermatologists on clinical intention, factuality, and specificity.

Our results reveal three main findings. First, we identify a substantial benchmark-to-bedside performance gap: although MLLMs demonstrate moderate diagnostic accuracy on public dermatology benchmarks, performance declines markedly when applied to real-world consultation cases. Second, incorporating referring clinicians' consultation notes improves diagnostic accuracy but also introduces substantial sensitivity to contextual information, with models frequently over-relying on incomplete or misleading notes, raising safety concerns for clinical deployment. Third, in contrast to recent work emphasizing reasoning capabilities, expert dermatologist review indicates that most diagnostic failures arise from incomplete or inaccurate extraction of key visual dermatologic features rather than deficiencies in downstream reasoning. [14,16,17] Together, these findings argue that realistic evaluation of dermatology MLLMs should prioritize clinical workflow fidelity, robustness to misleading context, and accurate visual grounding. We further provide practical recommendations for the development and responsible adoption of MLLMs in dermatology.

## RESULTS

### *Evaluation overview*

We evaluated model performance across three settings: public dermatology benchmarks, a real-world hospital dermatology consultation cohort, and targeted analyses of clinical context integration, severity-based triage, and reasoning behavior. Four image cohorts were used in this study (Supplementary Table 1). Collectively, these datasets comprise 92,947 clinical images from 46,980 cases spanning three public dermatology benchmarks and one real-world consult cohort. Public datasets included DermNet (19,559 images),[22] Fitzpatrick17k (16,577 images, Fitz17k)[23] and SCIN (10,406 images from 5,033 cases).[24] In addition, we evaluated a hospital consult dataset of 46,405 images from 5,811 annotated consult cases collected between 2014 and 2023, of which 2,351 are paired with clinical notes. Severe case frequency (assigned by LLM-as-a-judge with physician verification, see Methods) was similar across public datasets, ranging from 18.5% to 19.6%, but was substantially higher in the hospital consult dataset (44.0%). This indicates that the internal cohort was enriched for higher-acuity cases compared with public benchmarks.

### *Diagnosis performance on public datasets*

We first established baseline performance on standard public dermatology benchmarks under conventional evaluation settings. Zero-shot diagnostic performance was first assessed across three publicly available dermatology datasets: DermNet, Fitz17k, and SCIN. We used two diagnosis accuracy assessment methods established in the literature[31,32] (see Methods): a BERTScore scoring approach[33] and an LLM-as-a-judge[34] approach used to detect if the correct diagnosis was among the top-3 diagnoses output by the model for a random subset of each dataset. Overall, GPT-4.1 achieved the highest top-3 accuracy on all three public datasets, with accuracies of 42.25% on DermNet, 32.35% on Fitz17k, and 48.60% on SCIN (Figure 2). Among open-weight models, MedGemma-4B was the highest-performing model across datasets, achieving top-3 accuracies of 26.55% on DermNet, 20.00% on Fitz17k, and 39.50% on SCIN. These results for open-source models were consistent with BERTScore semantic similarity metric scoring as well (Supplementary Table 2). LLM-as-a-judge scoring was validated on a subset of cases independently reviewed by three board-certified or board-eligible dermatologists (Table 5).

*Diagnosis performance on real-world dataset without clinical context*

Applying the same models to real-world consultation cases revealed a substantial benchmark-to-bedside performance gap. Diagnostic accuracy declined markedly across almost all architectures relative to public benchmarks (with no architecture showing improved performance), despite using the same differential-diagnosis evaluation framework (Figure 2). On image-only consult cases, MedGemma-4B was the strongest-performing open-weight model, achieving the highest BERT-F1 score among open models (0.095) and the highest open-weight top-3 diagnostic accuracy (13.35%) (Supplementary Table 2). GPT-4.1 achieved the highest overall top-3 accuracy at 24.65%. Performance across almost all models dropped markedly compared to public benchmarks, highlighting the greater difficulty of real-world dermatology consultation cases. These trends were broadly consistent across BERTScore and LLM-as-a-judge analyses.

*Performance effects of clinical context incorporation on real-world dataset*

We next examined whether incorporating referring clinician consultation notes improves performance in real-world dermatology consultations. Clinical context in the form of the "the clinical question" was available for a significant fraction of the cases in the dataset and was extracted either through direct access to paired clinical consult notes referred to as "full EHR context" or generated through an "abbreviated" version of the question parsed by a dermatologist during case logging (Figure 2).

Compared with image-only prompting, the addition of clinical context substantially improved top-3 accuracy for InternVL, LLaVA-Med, MedGemma-4B, and GPT-4.1, whereas SkinGPT-4 improved only modestly (Supplementary Table 3). Using the note-derived context, MedGemma-4B again achieved the strongest semantic performance among open-weight models, with a BERT-F1 of 0.129, along with the highest open-weight top-3 accuracy at 28.75%, narrowly exceeding LLaVA-Med at 27.45%. GPT-4.1 achieved the highest overall performance with EHR context, with a BERT-F1 of 0.140 on the evaluated subset and a top-3 accuracy of 38.93%. Similar findings were made when using the abbreviated context from the consultant dermatologist (Supplementary Table 4). Clinician evaluation further supported the relative strength of adding clinical context, with MedGemma-4B and LLaVA-Med performing best among the open-weight models under clinical-context conditions (Table 5).

*Evaluation of enhancement/degradation from clinical context*

Although clinical context generally improved performance, models were also highly vulnerable to misleading or incomplete contextual information. To examine this more directly, we stratified consultation notes by the presence of leading diagnostic cues, defined as explicit differential diagnoses suggested by the referring clinician. Across all models, accuracy was higher with leading than non-leading context. With non-leading context, consult ("+EHR") top-3 accuracy ranged from 3.16% to 32.24%, compared with 9.46% to 46.61% with leading context (Table 1). Among open-weight models, LLaVA-Med improved from 18.97% to 36.85%, MedGemma-4B from 22.43% to 37.07%, and InternVL from 11.56% to 32.72%, whereas SkinGPT-4 remained low overall. GPT-4.1 achieved the highest performance in both settings, increasing from 32.24% to 46.61%. Even without leading cues, clinical context still improved performance relative to image-only evaluation across all models.

We then quantified enhancement and degradation after adding context relative to image-only inputs. With non-leading context, enhancement ranged from 2.9% for SkinGPT-4 to 21.5% for GPT-4.1, with intermediate gains for InternVL (10.5%), MedGemma-4B (16.7%), and LLaVA-Med (18.1%) (Table 1). Degradation in the non-leading setting was highest for SkinGPT-4 (84.2%) and InternVL (74.7%), remained substantial for LLaVA-Med (56.8%), and was lower for MedGemma-4B (38.9%) and GPT-4.1 (32.8%). With leading context, enhancement rose further, reaching 37.5% for GPT-4.1, 37.2% for LLaVA-Med, 33.6% for MedGemma-4B, and 32.0% for InternVL. However, degradation also remained substantial for several models, particularly LLaVA-Med (70.7%) and SkinGPT-4 (66.7%). MedGemma-4B achieved strong enhancement with lower degradation (42.0%); GPT-4.1 also showed both the highest enhancement and the lowest degradation (23.0%), indicating the greatest robustness to potentially misleading clinical information.

*Severity assessment performance for image-only and with clinical context*

Because exact diagnosis remained challenging, we next examined whether MLLMs could more reliably support severity-based triage of urgent dermatologic conditions. We evaluated both model outputs and gold-standard diagnoses using an LLM-as-a-judge approach. We also performed a manual dermatologist review of severity assignments in a subset of cases, with further analyses of concordance presented below. We focused on sensitivity for severe cases, given the importance of not missing high-acuity presentations.

On public datasets, no model achieved sensitivity above 0.50, and MedGemma-4B generally showed the strongest sensitivity among open-weight models, reaching 0.380 on DermNet, 0.454 on Fitz17k, and 0.478 on SCIN (Figure 3, Supplementary Table 5). On real-world consult data without clinical context, sensitivity also remained below 0.50 for all models. MedGemma-4B achieved the highest sensitivity among open-weight models for image-only consult cases at 0.456 (GPT-4.1 reached 0.364). These findings suggest that image-only triage performance remained limited even when diagnostic severity, rather than exact diagnosis, was the target.

Clinical context substantially improved severe-case sensitivity on the real-world hospital consult dataset. With clinical note-derived context, several model-context combinations exceeded 50% sensitivity, and abbreviated clinical context further improved performance in some cases (Supplementary Table 6-7). LLaVA-Med achieved the highest severe-case sensitivity with both full and abbreviated context, reaching approximately 0.65-0.72, while MedGemma-4B and GPT-4.1 also showed strong performance, in the 0.55-0.65 range. Thus, the strongest severe-case recall under contextual prompting was observed for LLaVA-Med. When cases paired with leading context (clinical context that offers a differential diagnosis) was removed, sensitivity generally declined across models (whereas removal of the same cases in the absence of context did not significantly affect sensitivity for any model) (Figure 3). Under full EHR context with cases containing leading cues removed, MedGemma-4B retained the highest sensitivities (Figure 3, Supplementary Table 6), while under abbreviated context, LLaVA-Med remained highest (Supplementary Table 7) although overall differences between these two models remained modest. Overall, these results suggest that clinically informative context can meaningfully improve severe-case recall, but robustness to misleading context may serve as an important differentiator across models.

Given that severe cases were more common in the real-world consult dataset than in public benchmarks, we also assessed performance based on severity status. When diagnostic accuracy was stratified by severity, most models performed worse on severe than non-severe cases (Supplementary Table 8). MedGemma-4B was again the exception, with comparable or slightly higher accuracy in severe versus non-severe cases, although confidence intervals overlapped. Overall, these results suggest that triage may be better aligned with current multimodal LLM capabilities than precise diagnosis, while robustness to severe disease and incomplete context remains an important differentiator across models.

*Assessment of reasoning by dermatologists*

To investigate performance differences across MLLM architectures, we examined physician-rated model reasoning which highlighted the importance of morphology specificity as a major discriminator in model quality. Given the substantial effort involved in manual dermatologist evaluation, we manually examined a subset of open-weight models to enable detailed and interpretable analysis of visual reasoning behavior, thereby providing insights to guide future model development.

We first found that LLM-as-a-judge closely matched clinician evaluations for both diagnosis accuracy and severity across public and real-world datasets, with diagnostic F1 scores of ~0.87–0.88 and severity F1 scores of ~0.88–0.92 (Supplementary Table 9), supporting the reliability of automated judging. We then assessed reasoning across three dermatologist-rated domains: intent alignment, factuality, and specificity. Inter-rater reliability was high overall and consistently above 0.5, with excellent agreement for diagnosis correctness and intent alignment and moderate-to-good agreement for specificity and factuality across datasets (Supplementary Tables 10–11).

Across both public and real-world datasets, LLaVA-Med and MedGemma showed the highest intent alignment and factuality, often near ceiling (Table 2), regardless of whether the final diagnosis was correct. In contrast, specificity was the clearest discriminator of correct versus incorrect responses. For example, MedGemma's specificity scores were substantially higher for correct than incorrect responses in both the public dataset (4.2 vs ~2.2) and the hospital real-world dataset (4.1–4.3 vs 1.1–1.2). This separation was more pronounced than for factuality or intent alignment and was consistent across models. Clinical context did not substantially change intent alignment or factuality across correctness strata. Instead, the ability to identify and prioritize visually discriminative features best distinguished correct from incorrect predictions. Together, these findings suggest that correct visual feature recognition and interpretation, rather than general medical knowledge or prompt adherence, is the key reasoning factor underlying correct dermatologic diagnosis in multimodal LLMs.

## DISCUSSION

### *Main study conclusions*

This study demonstrates that benchmark performance substantially overestimates the real-world clinical capability of current dermatology MLLMs. Across public dermatology benchmarks, diagnostic performance was broadly consistent with prior reports; [15–17,35] however, when the same models were applied to real-world consultation cases, performance declined markedly across architectures. This benchmark-to-bedside gap likely reflects at least three challenges that are underrepresented in public datasets for training: a case mix enriched for severe and clinically complex diagnoses (particularly drug eruptions); cases involving wide open-ended differential diagnoses; and real hospital photography characterized by variable lighting, inconsistent framing, multiple lesions, and background distractors. Together, these findings challenge the generalizability of current benchmark evaluations and underscore the need for assessment frameworks that better reflect realistic consultation workflows.

A second major finding is that clinical context is both necessary for improving performance and a major source of diagnostic vulnerability. Interestingly, incorporating referring clinicians' consultation notes (often referred to as "the clinical question" in consultative medicine) improved diagnostic accuracy across models, confirming the importance of text-based clinical context for meaningful performance. However, context integration also introduced vulnerabilities: some models were sensitive to incomplete or misleading contexts that offered incorrect differential diagnoses. The proportion of cases in which adding context worsened performance was substantial across models, reaching as high as 84% for SkinGPT-4 and 75% for InternVL. GPT-4.1 was relatively robust, while open-weight models showed greater vulnerability to misleading context. This fragility has direct safety implications. In routine dermatology consultations, referring clinicians may provide brief supplemental notes that are often imprecise, incomplete, or even diagnostically incorrect; a model that over-anchors on such notes may propagate rather than correct diagnostic error. From a deployment perspective, clinical context should therefore be viewed as both informative and potentially harmful when misleading.

A third major finding is that diagnostic failure in current dermatology MLLMs appears to arise primarily from insufficient visual feature extraction rather than downstream reasoning deficits. In contrast to recent work attributing MLLM errors primarily to reasoning deficits,[16,17,35] expert dermatologist review identified specificity of visual feature extraction—not downstream reasoning or general medical knowledge—as the principal discriminator between correct and incorrect responses. Models that correctly diagnosed a condition consistently described morphologically precise and diagnosis-relevant features, whereas failing responses offered generic or non-specific visual descriptions despite otherwise factual language. These findings suggest that the principal bottleneck in current dermatology MLLMs lies in visual grounding rather than reasoning, and that advances in reasoning alone are unlikely to close the benchmark-to-bedside gap without corresponding improvements in dermatology-specific visual feature extraction.

*Distinguishing features of this study*

This study advances prior dermatology MLLM related studies by introducing greater clinical realism in both data and task design. First, from a data perspective, most prior studies of dermatology MLLMs have been evaluated on standard public benchmarks consisting of image-only datasets, often dominated by dermoscopic images.[15–17] Such benchmarks do not reflect real-world dermatology practice. More broadly, prior reviews of LLM evaluation in medicine have also highlighted the limited use of real-world clinical datasets and the need for more clinically grounded assessment.[36] In contrast, our study evaluates MLLMs in a real-world dermatology consultation cohort that includes paired clinical photographs and referring clinician consultation notes. This cohort is also enriched for severe diagnoses—including drug eruptions and other high-acuity presentations—relative to publicly available datasets, better reflecting the case mix encountered in hospital dermatology practice.

Second, from an application perspective, our evaluation is designed to mirror actual consultation workflow. Rather than focusing only on image-based diagnosis or framing diagnosis as a multiple-choice task, we incorporate clinical context from referring clinicians and assess both open-ended differential diagnosis generation and severity-based triage. This enables a direct assessment of how EHR-derived clinical context modulates MLLM performance in a realistic dermatology consultation setting.

*Clinical implications and recommendations*

Current MLLM performance remains insufficient for autonomous diagnostic use in consultative dermatology. Even under the best-performing conditions, top-3 diagnostic accuracy remained limited, reaching 38.93% for GPT-4.1 with clinical context and 28.75% for the strongest open-weight model, while image-only performance was substantially lower. Severity-based triage appeared more promising than open-ended differential diagnosis, suggesting that triage may represent a more realistic near-term use case. In particular, MLLMs may have value as screening adjuncts to flag potentially high-acuity presentations for expedited review, provided that clinical context is available and of sufficient quality.

However, even for triage, performance remained highly dependent on context quality and therefore does not support deployment without human oversight. The sensitivity of triage performance to contextual information, together with the consistent underperformance on severe cases relative to non-severe cases across most models, argues against clinical use in settings where high-acuity disease is plausible. MedGemma-4B showed a notable exception, with comparable or slightly higher accuracy on severe versus non-severe cases, but this observation requires further validation in prospective settings.

A central practical implication is that clinical context should be treated as both a performance enabler and a potential source of error. Deployment frameworks for dermatology MLLMs should therefore incorporate mechanisms to assess consultation note quality, detect potentially leading or misleading contextual cues, and flag cases in which model outputs appear overly anchored to clinician-provided impressions. Structured consultation templates that standardize the information provided by referring clinicians may reduce this variability and improve model reliability.

More fundamentally, our findings suggest that future progress in dermatology MLLMs will depend less on improved reasoning alone and more on stronger visual grounding under realistic clinical conditions. Future model development should therefore prioritize dermatology-specific visual representation learning, robustness to incomplete or misleading context, and evaluation frameworks that mirror actual consultation workflows rather than benchmark performance alone.

*Limitations*

Several limitations should be considered when interpreting these findings. First, although we aimed to mirror real-world dermatology consultation workflows, the consultation images were drawn from an inpatient dermatology service, and a proportion were captured by dermatology team members rather than referring providers. As a result, image quality may be higher than would typically be encountered in true primary care or emergency department referrals. Second, this study focused on two core clinical tasks—open-ended differential diagnosis generation and severity-based triage—which are central to dermatology consultation workflows. However, other important downstream applications, such as management recommendations, treatment planning, and follow-up guidance, were not evaluated and warrant future study. Third, full EHR context—including prior notes, laboratory results, and medication lists—was not incorporated, only the clinical context supplied directly by the referring clinician. This design choice better reflects certain consultation settings, particularly external referrals; however, future work could assess how model performance changes as additional structured and unstructured clinical context is incorporated. Fourth, given the rapid evolution of MLLMs, we evaluated five representative models spanning general-domain, medical-domain, and dermatology-oriented systems. Although these models capture a range of current approaches, they do not exhaust the expanding landscape of multimodal models. Continued benchmarking of newer architectures will be important as the field evolves.

## CONCLUSIONS

Current MLLM performance is insufficient for autonomous diagnostic use in consultative dermatology. Benchmark performance substantially overestimates real-world clinical capability, and even the strongest models showed limited diagnostic accuracy under realistic consultation conditions. Severity-based triage may represent a more feasible near-term application than open-ended diagnosis, but performance remains too dependent on clinical context quality to support deployment without human oversight. More fundamentally, our findings indicate that the major bottleneck in current dermatology MLLMs lies in fragile context integration and insufficient visual grounding rather than reasoning alone. Future development should therefore prioritize realistic workflow-based evaluation, robustness to misleading context, and stronger dermatology-specific visual representation learning.

## METHODS

We evaluated five MLLMs across public dermatology image benchmarks and a real-world hospital dermatology cohort. Models were assessed on diagnostic performance using structured prompt-based inference with image-only and image-plus-clinical-context settings. Full Methods, including dataset curation, prompt details, adjudication procedures, severity definitions, physician evaluation, and statistical analysis, are provided in the Supplementary Methods.

### *Datasets*

We assembled an evaluation set comprising three public dermatology image datasets and one internal consultative dermatology dataset. Public datasets included DermNet, Fitzpatrick17k (Fitz17k), and SCIN. Datasets were restricted to gross clinical dermatology photographs, excluding dermoscopy, and dermatopathology.[22–24] The internal dataset consisted of inpatient dermatology consult images collected at a single institution (under Yale institutional review), with all cases paired with brief dermatologist written logs and a subset linked directly to paired consult note text from the EHR. Internal images and associated text were manually screened for relevance and redacted for protected health information; leading clinical context was separately annotated manually.

### *Models and Evaluation*

We evaluated four open-source multimodal LLMs and one proprietary comparator using standardized prompts requesting either a most likely diagnosis or a three-item differential diagnosis, with clinical context included when applicable. Performance in terms of diagnostic accuracy was assessed using two complementary approaches: BERTScore-based comparison of extracted differential diagnoses against reference differentials (for this analysis, GPT-4.1 MLLM responses were only generated for a subset of each dataset due to API constraints; performance comparisons should be interpreted accordingly), and an LLM-as-a-judge pipeline used to extract diagnoses, assess synonym-level agreement with gold-standard labels, and assign severity against a predefined list of high-acuity dermatologic diagnoses.[31–34] A subset of model outputs was also evaluated by dermatologists for diagnostic agreement and response quality.

### *Statistics*

For continuous rating outcomes, we summarized group means with 95% confidence intervals. For proportions, including accuracy and severity-related metrics, 95% confidence intervals were calculated using Wilson score intervals. Additional details regarding confidence interval estimation, inter-rater reliability, and statistical thresholds are provided in the Supplementary Methods.

## FUNDING

This study was supported by the National Institutes of Health (NIH) National Library of Medicine under Award Number R01LM01460. This work was also supported by the National Institutes of Health under institutional training grant T32AR007016.

## CONTRIBUTIONS

Study design: RJ, HK, ZQ, JG, QC. Board-eligible or board-certified dermatologist evaluation of MLLM responses: RJ, MM, AH, JG. Analysis of data: RJ, HK, ZQ. Screening of clinical images for protected health information: RJ, ML, AS, SC. Parsing of clinical consult notes: XA. Dermatologist curation of clinical context: RJ. All authors reviewed the manuscript, approved the final version, and agreed to submit the manuscript for publication.

## ACKNOWLEDGEMENTS

We thank Dr. Lingfei Qian and Dr. Hua Xu for their assistance with HIPAA-compliant API access for commercial MLLMs.

## DATA SHARING

Analysis code and prompts used for MLLM evaluation will be publicly available in a GitHub repository at https://github.com/Yale-BIDS-Chen-Lab/dermatology-mllm-evaluation-2026. Consultation documentation with paired clinical images and dermatologist-authored logs containing abbreviated clinical context and diagnostic impressions will remain restricted and will not be publicly distributed.

## FIGURES

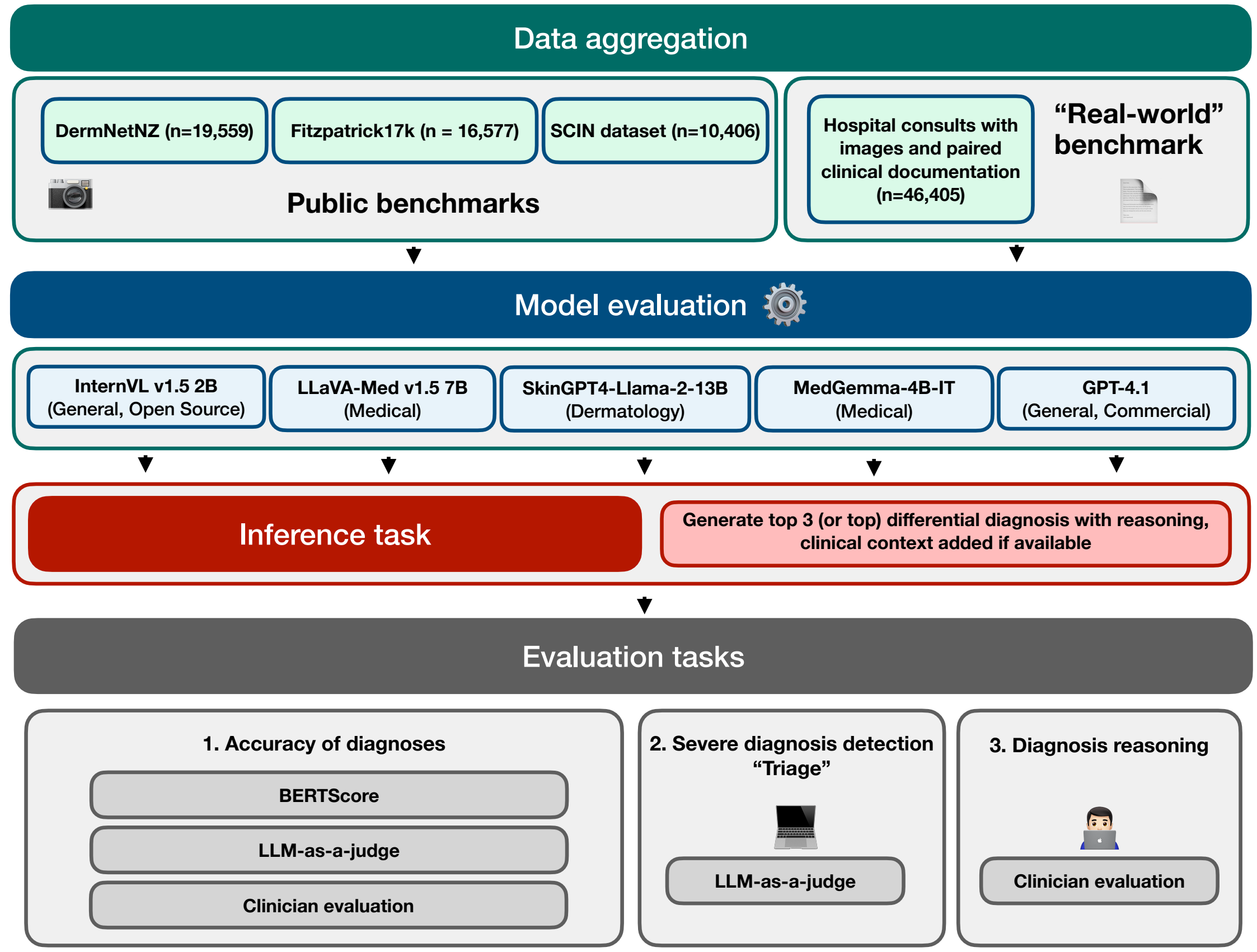

**Figure 1.** Study design for evaluation of multimodal large language models (MLLMs) in consultative dermatology. Public image-only datasets included DermnetNZ, Fitzpatrick17k, and the SCIN dataset. These were compared with a real-world retrospective inpatient dermatology consultation cohort consisting of paired clinical images and clinician-authored contextual narratives. Here, n refers to image count, not case count. Five MLLMs were evaluated: InternVL-Chat v1.5 (2B parameters, general domain), LLaVA-Med v1.5 (Mistral-7B backbone, medical domain), SkinGPT-4 (LLaMA v2 13B backbone, dermatology domain), MedGemma-4B (instruction-tuned, medical domain) and GPT-4.1 (commercial model, general domain). Models were tested under image-only and image-plus-context conditions. The principal task was differential diagnosis which was evaluated for accuracy and the correct identification of severe diagnoses for triage. Performance was assessed using BERTScore, LLM-as-a-judge and clinician evaluation.

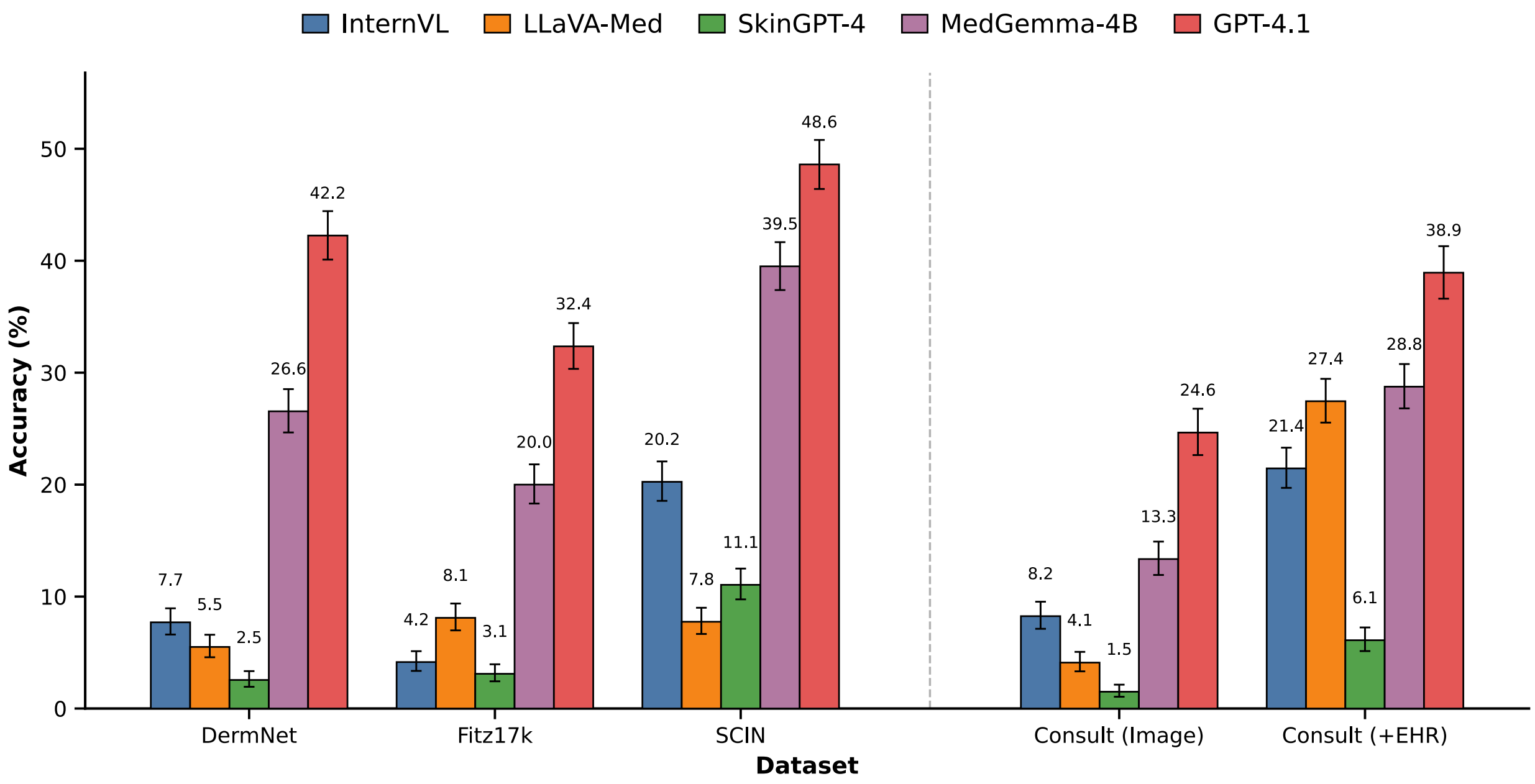

**Figure 2.** Top-3 diagnostic accuracy across datasets and models. Top-3 accuracy denotes the proportion of cases in which the reference diagnosis appeared among the top three model predictions. Accuracy was assessed using LLM-as-a-judge on a sampled subset of each dataset (n=2,000 per dataset); non-responses were excluded, and all valid responses were used in confidence interval calculations. Error bars indicate 95% Wilson score confidence intervals. Performance on the real-world hospital consult cohort (Consult) shown separately for image-only ("Image") and image plus clinical context (EHR-derived, "+EHR") input conditions.

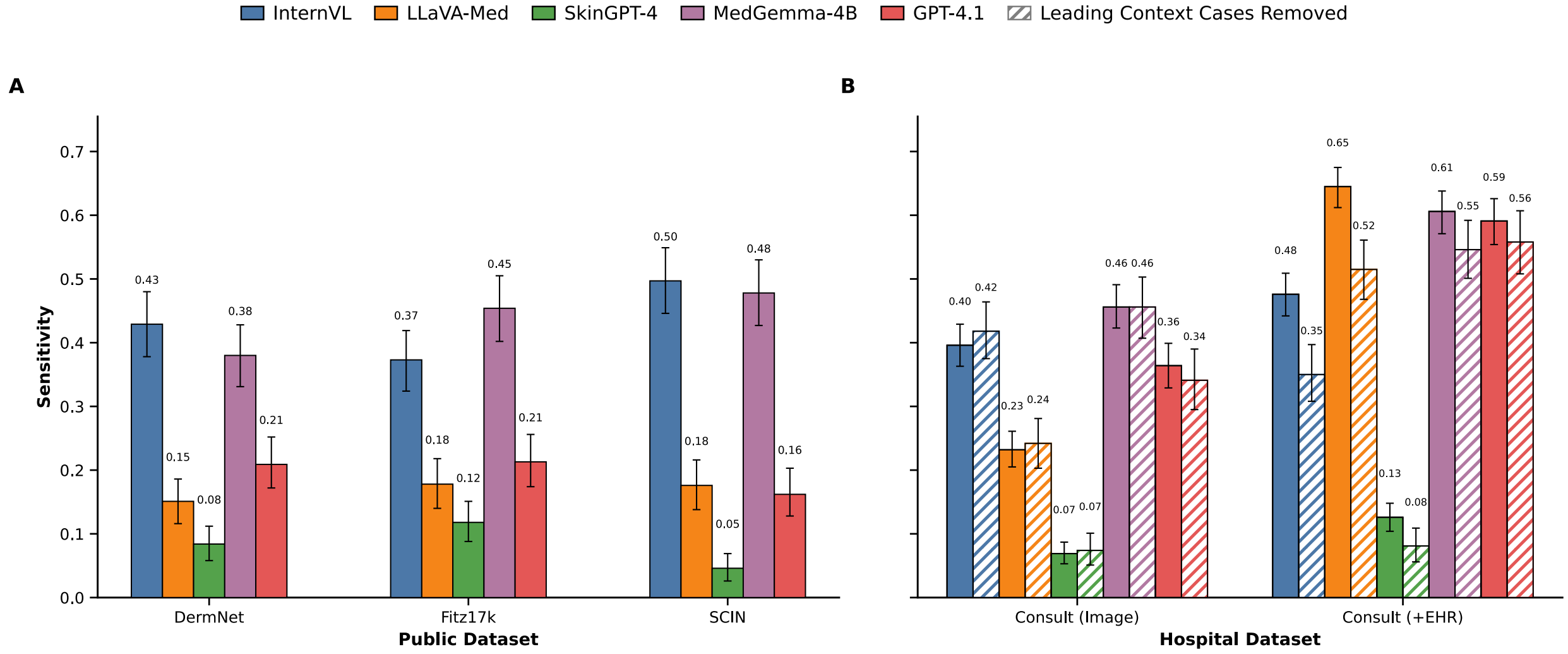

**Figure 3.** Sensitivity for severe diagnoses identification across public benchmark and hospital consult datasets by model. Panel A shows sensitivity on the public datasets DermNet, Fitz17k, and SCIN. Panel B shows sensitivity on the real-world hospital

consult dataset using image-only (“Image”) and image plus clinical context (EHR-derived, “+EHR”) input conditions. For real-world hospital consult datasets, leading context indicates whether clinical text preceding the image was retained or removed; removed-context bars are shown with hatched fill. Sensitivity was assigned by LLM-as-a-judge on a sampled subset of diagnoses from each dataset (n=2,000 per dataset). Non-responses were excluded, and all valid responses were used for confidence interval calculations. Error bars indicate 95% Wilson score confidence intervals.

## TABLES

| Model | Leading Context | Consult (Image) Accuracy (95% CI) | Consult (+EHR Context) Accuracy (95% CI) | Enhancement Rate (95% CI) | Degradation Rate (95% CI) |
|---|---|---|---|---|---|
| InternVL | No | 7.41% (5.95–9.19) | 11.56% (9.73–13.68) | 10.5% (8.7–12.6) | 74.7% (63.8–83.1) |
| InternVL | Yes | 7.39% (5.87–9.26) | 32.72% (29.76–35.82) | 32.0% (29.0–35.3) | 58.8% (47.0–69.7) |
| LLaVA-Med | No | 3.66% (2.66–5.00) | 18.97% (16.68–21.50) | 18.1% (15.8–20.6) | 56.8% (40.9–71.3) |
| LLaVA-Med | Yes | 4.46% (3.30–5.99) | 36.85% (33.79–40.01) | 37.2% (34.1–40.4) | 70.7% (55.5–82.4) |
| SkinGPT-4 | No | 1.88% (1.21–2.91) | 3.16% (2.25–4.43) | 2.9% (2.0–4.2) | 84.2% (62.4–94.5) |
| SkinGPT-4 | Yes | 0.98% (0.52–1.85) | 9.46% (7.73–11.52) | 9.2% (7.5–11.3) | 66.7% (35.4–87.9) |
| MedGemma-4B | No | 12.94% (11.02–15.15) | 22.43% (19.97–25.10) | 16.7% (14.4–19.3) | 38.9% (31.0–47.5) |
| MedGemma-4B | Yes | 14.24% (12.13–16.65) | 37.07% (34.00–40.23) | 33.6% (30.4–37.0) | 42.0% (33.9–50.5) |
| GPT-4.1 | No | 23.54% (20.81–26.50) | 32.24% (29.19–35.45) | 21.5% (18.5–24.8) | 32.8% (26.7–39.6) |
| GPT-4.1 | Yes | 23.11% (20.24–26.25) | 46.61% (43.08–50.18 | 37.5% (33.6–41.5) | 23.0% (17.4–29.8) |

**Table 1.** Effect of leading clinical context on internal consultation diagnostic accuracy. “Leading context” refers to clinical text that contains differential diagnosis suggestions preceding the image, compared to non-leading, descriptive context. "Consult (Image)" accuracy reflects top-3 diagnostic accuracy for image-only evaluation, while "Consult (+EHR Context)" accuracy reflects top-3 accuracy with accompanying free-text clinical information. Enhancement rate denotes the proportion of cases in which adding context improved diagnostic correctness relative to the image-only baseline, while degradation rate denotes the proportion in which accuracy worsened. Values are reported as percentages with 95% Wilson score confidence intervals.

| Model | Dataset | Accuracy (%) | Intent \| Correct | Intent \| Incorrect | Factuality \| Correct | Factuality \| Incorrect | Specificity \| Correct | Specificity \| Incorrect |
|---|---|---|---|---|---|---|---|---|
| InternVL | Public | 20.00% | 3.50 (2.05–4.95) | 3.38 (2.75–4.00) | 3.17 (2.38–3.96) | 2.83 (2.28–3.38) | 3.00 (1.85–4.15) | 1.87 (1.27–2.47) |
| LLaVA-Med | | 6.70% | 5.00 | 4.71 (4.31–5.00) | 4.00 | 4.33 (4.11–4.56) | 4.00 | 2.00 (1.61–2.39) |
| SkinGPT-4 | | 13.30% | 3.50 (1.91–5.00) | 2.19 (1.72–2.66) | 3.75 (2.23–5.00) | 2.54 (2.16–2.92) | 3.25 (1.73–4.77) | 1.73 (1.34–2.12) |
| MedGemma-4B | | 36.70% | 5.00 (5.00–5.00) | 4.89 (4.67–5.00) | 4.73 (4.41–5.00) | 4.16 (3.79–4.53) | 4.18 (3.59–4.77) | 2.21 (1.71–2.71) |
| InternVL | Consult (Image) | 6.70% | 4.00 | 2.71 (2.17–3.26) | 4.00 | 2.19 (1.81–2.56) | 3.50 | 1.11 (1.00–1.24) |
| LLaVA-Med | | 10.00% | 5.00 (5.00–5.00) | 4.78 (4.44–5.00) | 4.00 (4.00–4.00) | 4.00 (4.00–4.00) | 3.00 | 1.00 |
| SkinGPT-4 | | 3.30% | 2.00 | 1.24 (1.08–1.41) | 3.00 | 1.55 (1.33–1.77) | 3.00 | 1.03 (1.00–1.11) |
| MedGemma-4B | | 26.70% | 5.00 (5.00–5.00) | 4.64 (4.26–5.00) | 4.50 (4.05–4.95) | 3.36 (3.07–3.66) | 4.25 (3.86–4.64) | 1.23 (1.04–1.42) |
| InternVL | Consult (+EHR) | 26.70% | 3.62 (2.74–4.51) | 3.09 (2.50–3.69) | 3.00 (2.23–3.77) | 2.32 (1.88–2.76) | 3.00 (2.00–4.00) | 1.18 (1.01–1.36) |
| LLaVA-Med | | 30.00% | 5.00 (5.00–5.00) | 4.62 (4.18–5.00) | 4.00 (3.59–4.41) | 3.59 (3.31–3.87) | 2.67 (2.22–3.11) | 1.00 (1.00–1.00) |
| SkinGPT-4 | | 3.30% | 2.00 | 1.48 (1.24–1.72) | 3.00 | 1.83 (1.60–2.06) | 3.00 | 1.03 (1.00–1.11) |
| MedGemma-4B | | 33.30% | 4.80 (4.35–5.00) | 4.90 (4.76–5.00) | 4.40 (3.90–4.90) | 3.80 (3.51–4.09) | 4.10 (3.57–4.63) | 1.15 (1.00–1.32) |

**Table 2.** Physician-rated diagnostic quality stratified by dataset type. Clinician evaluation of model responses across three domains: intent alignment (whether the response addresses the clinical task), factuality (accuracy of medical statements), and specificity (degree of diagnosis-relevant morphology detail). Each domain was rated on a 5-point Likert scale (1 = poor, 5 = excellent) by board-certified dermatologists. Higher scores indicate better performance in each domain. Ratings are shown separately for responses with correct versus incorrect diagnoses (based on top-3 accuracy) to assess qualitative differences independent of final diagnostic outcome. Values represent mean Likert scores with 95% confidence intervals.

# Supplementary Material

## SUPPLEMENTAL FIGURES AND TABLES

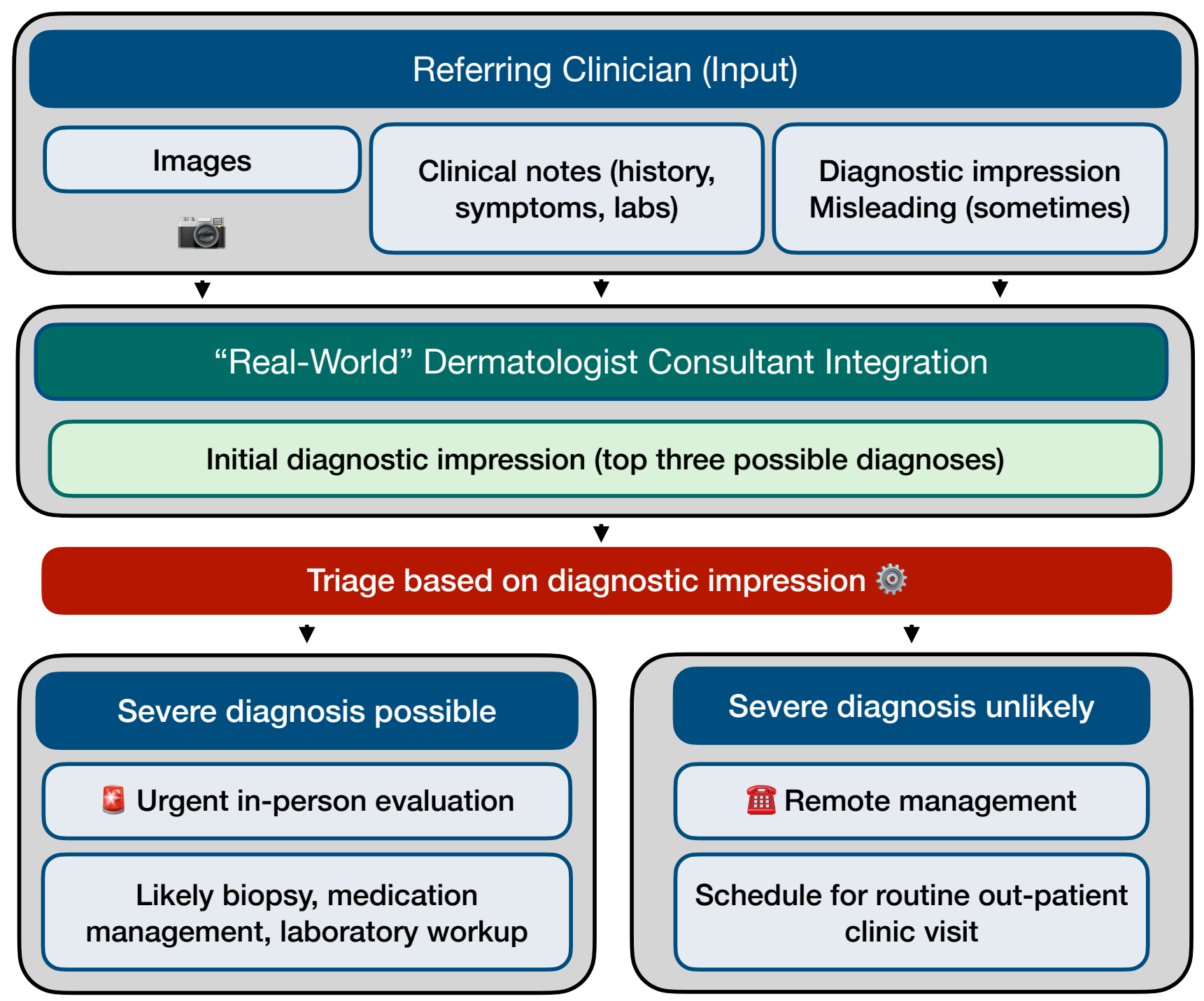

**Figure S1.** Real-world dermatology workflow showing principal steps in enabling appropriate differential diagnosis-based triage of cases from referring clinicians. Major steps are highlighted including the provision of clinical context by the referring clinician, the generation of a differential diagnosis by the dermatologist and triage of the case based on the range of potential differential diagnoses.

**Table S1. Summary of datasets used in evaluation**

| Dataset | Total Images | Total Cases | Source | Type | Severe Case Frequency |
|---|---|---|---|---|---|
| DermNet | 19,559 | 19,559 | DermNet NZ Trust [16] | Public | 0.1960 |
| Fitz17k | 16,577 | 16,577 | Groh et al., CVPRW 2021 [17] | Public | 0.1905 |
| SCIN | 10,406 | 5,033 | Ward et al., JAMA Netw Open 2024 [22] | Public | 0.1850 |
| Consult | 46,405 | 5,811 (2,351 with paired clinical note) | Yale New Haven Health dermatology consults (2014–2023) | Internal | 0.4400 |

**Table S1.** Summary of datasets used in evaluation. Shown are the total numbers of images and cases, data source, dataset type, and severe case frequency for each dataset (assigned by LLM-as-a-judge with manual verification). Severe case frequency denotes the proportion of cases labeled as severe within each dataset computed from a sampled subset. For DermNet and Fitzpatrick17k, each image is treated as one case. The "Consult" dataset consisted of YNHH dermatology consults collected from 2014 to 2023, including a subset of cases with paired clinical notes.

**Table S2. Diagnostic performance across public datasets for differential diagnosis**

| Dataset | Model | BERT-F1 | BERT-Precision | BERT-Recall | Top-3 Accuracy (95% CI) |
|---|---|---|---|---|---|
| DermNet | InternVL | 0.128 (0.126-0.129) | 0.217 (0.214-0.220) | 0.121 (0.119-0.123) | 7.70% (6.61–8.95%) |
| | LLaVA-Med | 0.144 (0.143-0.146) | 0.205 (0.202-0.207) | 0.193 (0.191-0.195) | 5.50% (4.58–6.59%) |
| | SkinGPT-4 | 0.126 (0.125-0.128) | 0.198 (0.195-0.201) | 0.151 (0.149-0.153) | 2.55% (1.94–3.34%) |
| | MedGemma-4B | 0.174 (0.172-0.176) | 0.331 (0.327-0.334) | 0.144 (0.143-0.146) | 26.55% (24.66–28.53%) |
| | GPT-4.1 | 0.194 (0.186-0.203) | 0.393 (0.377-0.409) | 0.151 (0.143-0.159) | 42.25% (40.10%–44.43%) |
| Fitz17k | InternVL | 0.112 (0.111-0.114) | 0.197 (0.194-0.200) | 0.112 (0.110-0.114) | 4.15% (3.36–5.12%) |
| | LLaVA-Med | 0.115 (0.114-0.117) | 0.158 (0.155-0.160) | 0.166 (0.164-0.169) | 8.10% (6.98–9.38%) |
| | SkinGPT-4 | 0.111 (0.109-0.113) | 0.165 (0.162-0.167) | 0.147 (0.144-0.149) | 3.10% (2.43–3.95%) |
| | MedGemma-4B | 0.148 (0.147-0.150) | 0.271 (0.267-0.274) | 0.134 (0.132-0.136) | 20.00% (18.31–21.81%) |
| | GPT-4.1 | 0.170 (0.161-0.178) | 0.335 (0.320-0.351) | 0.144 (0.135-0.153) | 32.35% (30.34%–34.43%) |
| SCIN | InternVL | 0.107 (0.105-0.109) | 0.104 (0.101-0.106) | 0.171 (0.168-0.174) | 20.25% (18.55–22.07%) |
| | LLaVA-Med | 0.064 (0.063-0.066) | 0.054 (0.052-0.055) | 0.149 (0.145-0.152) | 7.75% (6.66–9.00%) |
| | SkinGPT-4 | 0.120 (0.118-0.122) | 0.130 (0.127-0.133) | 0.212 (0.208-0.215) | 11.05% (9.75–12.50%) |
| | MedGemma-4B | 0.185 (0.182-0.188) | 0.175 (0.172-0.178) | 0.226 (0.222-0.230) | 39.50% (37.38–41.66%) |
| | GPT-4.1 | 0.206 (0.194-0.218) | 0.207 (0.195-0.218) | 0.226 (0.213-0.239) | 48.60% (46.41%–50.79%) |

BERTScore values are reported as F1 (F1), precision, and recall. Top-3 accuracy denotes the fraction of cases in which the reference diagnosis appeared among the top three model predictions. Top-3 accuracy was calculated using LLM-as-a-judge on a sample of the dataset (n=2000 for each dataset, non-responses were excluded and total responses were used for confidence interval calculations). Confidence intervals correspond to 95% Wilson score intervals.

**Table S3. Diagnostic performance across real-world datasets for differential diagnosis**

| Dataset | Model | BERT-F1 | BERT-Precision | BERT-Recall | Top-3 Accuracy (95% CI) |
|---|---|---|---|---|---|
| Consult (Image) | InternVL | 0.086 (0.085-0.086) | 0.130 (0.129-0.131) | 0.093 (0.092-0.094) | 8.25% (7.12–9.54%) |
| | LLaVA-Med | 0.083 (0.083-0.084) | 0.127 (0.126-0.128) | 0.099 (0.098-0.100) | 4.10% (3.32–5.06%) |
| | SkinGPT-4 | 0.053 (0.052-0.053) | 0.072 (0.071-0.073) | 0.075 (0.074-0.076) | 1.50% (1.05–2.13%) |
| | MedGemma-4B | 0.095 (0.094-0.096) | 0.151 (0.150-0.152) | 0.094 (0.093-0.095) | 13.35% (11.93–14.91%) |
| | GPT-4.1 | 0.095 (0.088-0.101) | 0.174 (0.163-0.186) | 0.084 (0.077-0.092) | 24.65% (22.64%–26.78% |
| Consult (+EHR) | InternVL | 0.102 (0.100-0.104) | 0.161 (0.158-0.164) | 0.106 (0.104-0.109) | 21.45% (19.71–23.30%) |
| | LLaVA-Med | 0.109 (0.107-0.111) | 0.186 (0.183-0.190) | 0.103 (0.100-0.105) | 27.45% (25.54–29.45%) |
| | SkinGPT-4 | 0.067 (0.066-0.069) | 0.091 (0.089-0.094) | 0.097 (0.094-0.099) | 6.10% (5.13–7.24%) |
| | MedGemma-4B | 0.129 (0.126-0.131) | 0.201 (0.198-0.205) | 0.127 (0.124-0.130) | 28.75% (26.81–30.77%) |
| | GPT-4.1 | 0.140 (0.130-0.149) | 0.233 (0.218-0.248) | 0.132 (0.120-0.143) | 38.93% (36.61%–41.30%) |

BERTScore values are reported as F1 (F1), precision, and recall. Image-only queries used clinical photographs without accompanying clinical context. EHR queries incorporated free-text clinical context from consultation notes. Top-3 accuracy denotes inclusion of the reference diagnosis among the top three predictions. Top-3 accuracy was calculated using LLM-as-a-judge on a sample of the dataset (n=2000 for each dataset, non-responses were excluded and total responses were used for confidence interval calculations). Confidence intervals correspond to 95% Wilson score intervals.

**Table S4. Abbreviated-context performance on hospital consultation dataset**

| Dataset | Model | BERTScore-F1 | BERTScore-Precision | BERTScore-Recall | Top-3 Accuracy (95% CI) |
|---|---|---|---|---|---|
| Consult (+Abbreviated) | InternVL | 0.111 (0.110-0.113) | 0.173 (0.171-0.175) | 0.116 (0.114-0.117) | 25.10% (23.25–27.05%) |
| | LLaVA-Med | 0.118 (0.116-0.120) | 0.202 (0.200-0.205) | 0.111 (0.109-0.112) | 33.25% (31.22–35.35%) |
| | SkinGPT-4 | 0.071 (0.070-0.072) | 0.095 (0.093-0.097) | 0.102 (0.100-0.103) | 7.30% (6.24–8.52%) |
| | MedGemma-4B | 0.129 (0.128-0.131) | 0.197 (0.195-0.200) | 0.133 (0.131-0.135) | 30.85% (28.86–32.91%) |
| | GPT-4.1 | 0.137 (0.128-0.145) | 0.235 (0.222-0.248) | 0.127 (0.118-0.136) | 43.15% (40.79%–45.54%) |

BERTScore values are reported as F1 (F1), precision, and recall and computed for the entire dataset except for GPT-4.1 which was only run on a sub-sample of the dataset (corresponding to the sub-sample used for LLM-as-a-judge evaluations). Top-3 accuracy denotes the fraction of cases in which the reference diagnosis appeared among the top three model predictions. Top-3 accuracy was calculated using LLM-as-a-judge on a sample of the dataset (n=2000 for each dataset, non-responses were excluded and total responses were used for confidence interval calculations). Confidence intervals correspond to 95% Wilson score intervals.

**Table S5. Sensitivity and specificity for detecting severe diagnoses for public datasets**

| Dataset | Model | Sensitivity | Specificity |
|---|---|---|---|

| DermNet | InternVL | 0.429 (0.378-0.480) | 0.593 (0.568-0.618) |
|---|---|---|---|
| | LLaVA-Med | 0.151 (0.116-0.186) | 0.779 (0.759-0.799) |
| | SkinGPT-4 | 0.084 (0.058-0.112) | 0.932 (0.918-0.944) |
| | MedGemma-4B | 0.380 (0.331-0.428) | 0.756 (0.736-0.776) |
| | GPT-4.1 | 0.209 (0.172-0.252) | 0.916 (0.901-0.929) |
| Fitz17k | InternVL | 0.373 (0.324-0.419) | 0.607 (0.582-0.631) |
| | LLaVA-Med | 0.178 (0.140-0.218) | 0.851 (0.834-0.867) |
| | SkinGPT-4 | 0.118 (0.088-0.151) | 0.907 (0.894-0.921) |
| | MedGemma-4B | 0.454 (0.402-0.505) | 0.771 (0.750-0.791) |
| | GPT-4.1 | 0.213 (0.174-0.256) | 0.897 (0.882-0.911) |
| SCIN | InternVL | 0.497 (0.446-0.549) | 0.499 (0.477-0.524) |
| | LLaVA-Med | 0.176 (0.138-0.216) | 0.828 (0.809-0.846) |
| | SkinGPT-4 | 0.046 (0.026-0.069) | 0.963 (0.954-0.972) |
| | MedGemma-4B | 0.478 (0.427-0.530) | 0.633 (0.611-0.656) |
| | GPT-4.1 | 0.162 (0.128-0.203) | 0.918 (0.903-0.930) |

Severity frequency as well sensitivity and specificity is assigned by LLM-as-a-judge from a sample of diagnoses (n=2000 for each dataset, non-responses were excluded and total responses were used for confidence interval calculations). Confidence intervals correspond to 95% Wilson score intervals.

**Table S6. Sensitivity and specificity for detecting severe diagnoses for hospital datasets**

| **Dataset** | **Model** | **Sensitivity** | **Specificity** | **Leading Context** |
|---|---|---|---|---|
| Consult (Image) | InternVL | 0.396 (0.363-0.429) | 0.614 (0.585-0.643) | Not removed |
| | LLaVA-Med | 0.232 (0.205-0.261) | 0.747 (0.721-0.773) | Not removed |
| | SkinGPT-4 | 0.069 (0.053-0.087) | 0.945 (0.931-0.958) | Not removed |
| | MedGemma-4B | 0.456 (0.423-0.491) | 0.611 (0.582-0.640) | Not removed |
| | GPT-4.1 | 0.364 (0.329-0.399) | 0.732 (0.703-0.759) | Not removed |
| | InternVL | 0.418 (0.375-0.464) | 0.603 (0.563-0.642) | Removed |
| | LLaVA-Med | 0.242 (0.203-0.281) | 0.742 (0.705-0.778) | Removed |
| | SkinGPT-4 | 0.074 (0.051-0.101) | 0.958 (0.941-0.973) | Removed |
| | MedGemma-4B | 0.456 (0.407-0.503) | 0.620 (0.582-0.660) | Removed |
| | GPT-4.1 | 0.341 (0.295-0.390) | 0.763 (0.722-0.799) | Removed |
| Consult (+EHR) | InternVL | 0.476 (0.442-0.509) | 0.635 (0.607-0.663) | Not removed |

| | LLaVA-Med | 0.645 (0.612-0.675) | 0.492 (0.461-0.522) | Not removed |
|---|---|---|---|---|
| | SkinGPT-4 | 0.126 (0.104-0.148) | 0.882 (0.862-0.901) | Not removed |
| | MedGemma-4B | 0.606 (0.571-0.638) | 0.519 (0.490-0.549) | Not removed |
| | GPT-4.1 | 0.591 (0.554-0.626) | 0.595 (0.564-0.626) | Not removed |
| | InternVL | 0.350 (0.308-0.397) | 0.714 (0.677-0.750) | Removed |
| | LLaVA-Med | 0.515 (0.468-0.561) | 0.520 (0.479-0.560) | Removed |
| | SkinGPT-4 | 0.081 (0.056-0.109) | 0.938 (0.917-0.957) | Removed |
| | MedGemma-4B | 0.546 (0.501-0.592) | 0.533 (0.491-0.574) | Removed |
| | GPT-4.1 | 0.558 (0.508-0.607) | 0.621 (0.577-0.664) | Removed |

Leading context indicates whether clinical text preceding the image was retained or removed. Severity frequency as well sensitivity and specificity is quantified from severity assigned by LLM-as-a-judge from a sample of diagnoses (n=2000 for each dataset, non-responses were excluded and total responses were used for confidence interval calculations). Confidence intervals correspond to 95% Wilson score intervals.

**Table S7. Severity under abbreviated clinical context**

| Dataset | Model | Sensitivity | Specificity | Leading Context |
|---|---|---|---|---|
| Consult (+Abbreviated) | InternVL | 0.551 (0.518-0.582) | 0.626 (0.596-0.655) | Not removed |
| | LLaVA-Med | 0.721 (0.690-0.752) | 0.483 (0.453-0.512) | Not removed |
| | SkinGPT-4 | 0.167 (0.142-0.193) | 0.919 (0.902-0.934) | Not removed |
| | MedGemma-4B | 0.649 (0.618-0.681) | 0.574 (0.545-0.603) | Not removed |
| | GPT-4.1 | 0.604 (0.568-0.639) | 0.654 (0.623-0.684) | Not removed |
| | InternVL | 0.501 (0.456-0.549) | 0.617 (0.577-0.656) | Removed |
| | LLaVA-Med | 0.677 (0.634-0.721) | 0.482 (0.440-0.524) | Removed |
| | SkinGPT-4 | 0.133 (0.104-0.166) | 0.931 (0.910-0.951) | Removed |
| | MedGemma-4B | 0.614 (0.570-0.660) | 0.561 (0.518-0.601) | Removed |

| | GPT-4.1 | 0.588 (0.538-0.637) | 0.681 (0.637-0.721) | Removed |
|---|---|---|---|---|

Leading context indicates whether clinical text preceding the image was retained or removed. Severity frequency as well sensitivity and specificity is assigned by LLM-as-a-judge from a sample of diagnoses (n=2000 for each dataset, non-responses were excluded and total responses were used for confidence interval calculations). Confidence intervals correspond to 95% Wilson score intervals.

**Table S8. Top-3 diagnostic accuracy stratified by disease severity on real-world dataset**

| **Model** | **Severity** | **Top-3 Accuracy (95% CI)** |
|---|---|---|
| InternVL | Not severe | 9.06% (7.49-10.92) |
| | Severe | 5.29% (3.98-7.01) |
| LLaVA-Med | Not severe | 4.34% (3.28-5.73) |
| | Severe | 3.65% (2.58-5.13) |
| SkinGPT-4 | Not severe | 2.22% (1.50-3.28) |
| | Severe | 0.47% (0.18-1.20) |
| MedGemma-4B | Not severe | 12.29% (10.47-14.38) |
| | Severe | 15.18% (12.92-17.74) |
| GPT-4.1 | Not severe | 28.62% (25.82–31.59) |
| | Severe | 19.50% (16.78–22.55) |

Accuracy reflects Top-3 diagnostic accuracy with 95% confidence intervals. Accuracy and severity is assigned by LLM-as-a-judge from a sample of diagnoses (n=2000 for each dataset, non-responses were excluded and total responses were used for confidence interval calculations). Confidence intervals correspond to 95% Wilson score intervals.

**Table S9. Accuracy and severity classification performance**

| **Dataset** | **Accuracy \| Precision** | **Accuracy \| Recall** | **Accuracy \| F1** | **Severity \| Precision** | **Severity \| Recall** | **Severity \| F1** | **Severity \| Sensitivity** | **Severity \| Specificity** |
|---|---|---|---|---|---|---|---|---|
| Consult | 0.9444 | 0.8095 | 0.8718 | 0.9429 | 0.8919 | 0.9167 | 0.8919 | 0.9901 |
| Public | 0.9500 | 0.8261 | 0.8837 | 0.8462 | 0.9167 | 0.8800 | 0.9167 | 0.9815 |

Concordance of precision, recall, F1, sensitivity, and specificity are reported for accuracy and severity binary classification tasks between dermatologist evaluations and LLM-as-a-judge evaluations. Dermatologist responses were used as the gold standard. Metrics were computed only for evaluations images with both LLM-as-a-judge and dermatologist responses.

**Table S10. Inter-rater reliability for public dataset responses among three experts**

| **Outcome** | **Scale** | **Count** | **ICC2** | **95% CI** |
|---|---|---|---|---|
| Diagnosis correctness | Binary | 60 | 0.936 | [0.9, 0.96] |
| Factuality | Likert | 44 | 0.576 | [0.4, 0.72] |
| Intent Alignment | Likert | 59 | 0.756 | [0.65, 0.84] |
| Specificity | Likert | 45 | 0.632 | [0.48, 0.76] |
| Severity Score | Likert | 60 | 0.679 | [0.56, 0.78] |

ICC2 values are calculated for dermatologist responses from a panel of 3-raters for either Binary or Likert scale outcomes. Only responses from raters who completed the rating task were assessed.

**Table S11. Inter-rater reliability for hospital consultation dataset responses among three experts**

| Outcome | Scale | Count | ICC2 | 95% CI |
|---|---|---|---|---|
| Diagnosis correctness | Binary | 20 | 0.835 | [0.69, 0.93] |
| Factuality | Likert | 16 | 0.818 | [0.64, 0.93] |
| Intent Alignment | Likert | 20 | 0.933 | [0.87, 0.97] |
| Specificity | Likert | 14 | 0.687 | [0.41, 0.87] |
| Severity Score | Likert | 20 | 0.828 | [0.66, 0.92] |

ICC2 values are calculated for dermatologist responses from a panel of 3-raters for either Binary or Likert scale outcomes. Only responses from raters who completed the rating task were assessed.

## SUPPLEMENTAL METHODS

### *Datasets*

We restricted our choice of public dermatology datasets to those exclusively containing gross clinical dermatology images, while removing any dermoscopy or images containing bedside diagnostics or dermatopathology:  DermNet, Fitzpatrick17k (Fitz17k) and SCIN dataset.[16,17] Datasets were selected on the basis of having a sufficient variety of clinical dermatologic diagnoses to reflect the range seen in both real world out-patient and in-patient dermatology settings. Metrics were quantified separately for each public dataset when possible, to account for variability in dataset diagnoses composition.

IRB exemption was obtained from Yale institutional review under #2000041054 to use archived clinical notes and images from a single institution related to consultative dermatology. Consultative dermatology images corresponding to those captured in the in-patient dermatology setting were accumulated from 2014 to 2024. These images were manually screened for relevance as well as the presence of visual PHI by at least two reviewers/co-authors. Flagged images were subsequently removed. Information including the hospital site, consulting provider, an abbreviated version of clinical context and presumed diagnosis or differential diagnosis upon consult completion were recorded. We retrieved OMOP-structured EHR data from patients who had seen been by the dermatology service in the hospital. We isolated initial consult notes to extract the original consult order text which was isolated from the notes using a series of regular expressions. This was matched with consultative images using medical record numbers. Any PHI contained in either the consult order text or the abbreviated clinical context was identified and redacted in a PHI-compliant manner.

Leading queries were also annotated manually in the same process. Leading queries were defined as those queries that propose a potentially dermatologic diagnosis. References to a prior history of a dermatologic diagnosis or medications/symptoms/medical history specific to a dermatologic diagnosis were not considered leading.

*MLLM selection and prompts*

We selected four open-source MLLM architectures and one commercial MLLM to reflect the diverse architectures commonly used in this field:

- InternVL-Chat v1.5 (2B parameters, “OpenGVLab/Mini-InternVL-Chat-2B-V1-5”) is a general-domain MLLM with jointly-trained ViT-derived visual and text tokens.
- LLaVA-Med v1.5 (Mistral-7B backbone, medical domain, “microsoft/llava-med-v1.5-mistral-7b”) is a widely used medical-domain MLLM, pretrained on several million medical image-text pairs, using a pretrained CLIP ViT vision encoder.
- SkinGPT4 is a publicly available dermatology-domain specific MLLM using a pretrained CLIP encoder on dermatology images with a Llama-2-13b backbone. We exclusively used the pretrained publicly available version.
- MedGemma-4B Instruct (“google/medgemma-4b-it”) is a medical-domain MLLM with both an image encoder and language encoder trained on medical domain tasks including dermatology.
- GPT-4.1 is a commercial general-domain MLLM from OpenAI that accepts text and image inputs and produces text outputs, with strong instruction-following and long-context capabilities; we used it as the proprietary comparator in our benchmark. GPT-4.1 was accessed through the hospital-managed Azure OpenAI environment (API version 2024-10-21).

The following prompts were used for MLLM response generation. The same prompt was used for all four MLLM architectures to maintain consistency.
- differential diagnosis: "Please offer the three most likely diagnoses for this dermatology image."
- differential diagnosis with clinical context: "Please offer the three most likely diagnoses with this context: {context}",

*Example clinical context*

An illustrative case is provided below with an image obtained from the hospital from a consult to dermatology (reviewed to be appropriate for dissemination). The referring practitioner provided clinical context as part of the consultative process. A gold standard diagnosis and/or differential is also provided corresponding to the case, authored by the consultant.

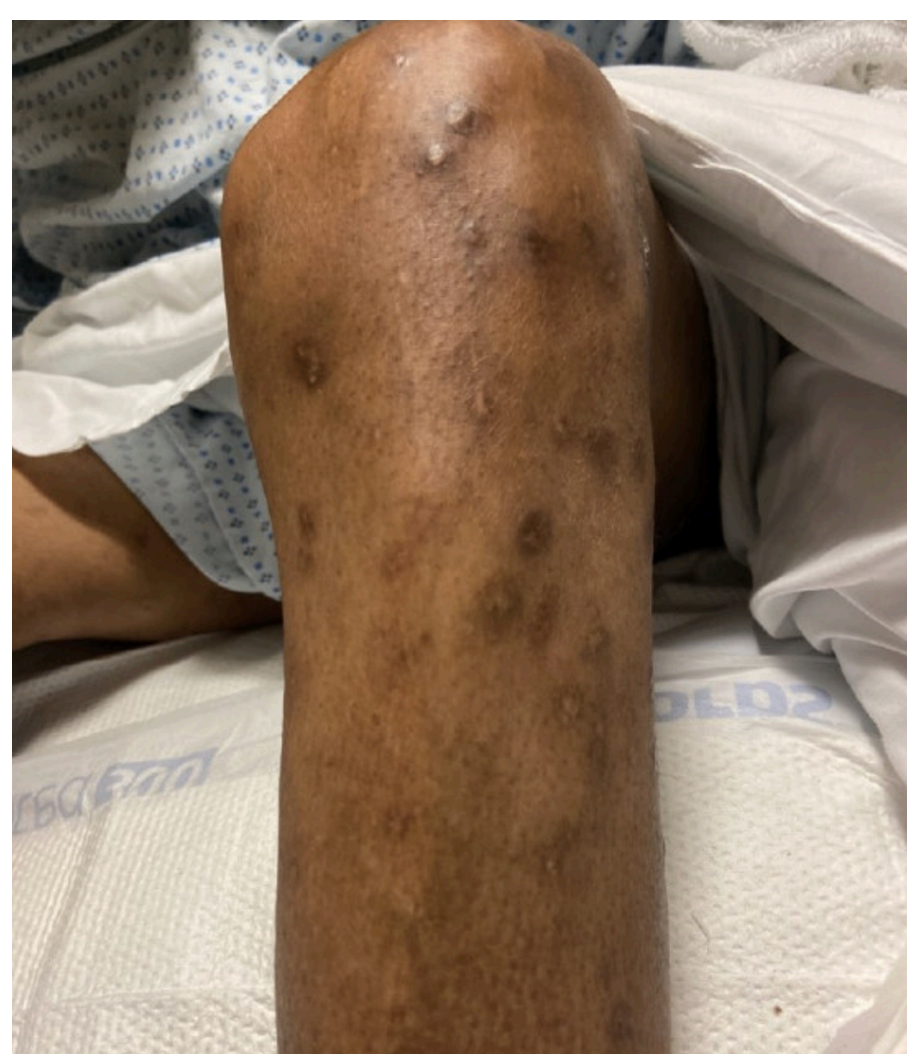

clinical context: “itchy bumpy rash BL UE and LE”
gold standard diagnosis: “prurigo nodularis/pruritus”

*Severity assessment*

Severity was defined based on a historic list of diagnosis corresponding to those that should be seen urgently or emergently for dermatology consultation based on the views of senior dermatologists. This definition may potentially differ from institution to institution however notable significant diagnoses such as SJS/TEN or erythrodermic psoriasis are generally considered severe. These lists were used for both LLM-as-a-judge and made available to clinician evaluation (dermatologist-as-judge) at the point of evaluation. This was performed with a Likert scale for how closely a diagnosis semantically matched a diagnosis in the list. Values of 1 were considered non-severe while values greater than 1 were considered severe. The complete list is provided here:

```{text}
Necrotizing fasciitis
Erysipelas or cellulitis (central facial)
Purpura fulminans
Neonatal herpes simplex
Genital herpes in antepartum female
Disseminated herpes zoster
Angioedema with tracheal obstruction
Hereditary angioneurotic edema — acute attack
Meningococcemia
Toxic epidermal necrolysis
Stevens-Johnson syndrome
Staphylococcal scalded skin syndrome
Staph/strep toxic shock syndrome
Drug hypersensitivity syndrome (including DRESS)
Rocky Mountain spotted fever
```

```
Infantile hemangioma with airway obstruction
Plague
Anthrax
Smallpox
Kawasaki disease
Brown recluse spider or black widow bite in infant
Ecthyma gangrenosum
Exfoliative erythroderma
Pustular psoriasis
Acute generalized exanthematous pustulosis (AGEP)
Gonococcemia
Kaposi's varicelliform eruption / eczema vaccinatum
Varicella / disseminated zoster
Acute systemic lupus erythematosus or dermatomyositis
Pemphigus vulgaris
Nodular or ulcerating melanoma
```

*BERTScore differential diagnosis matrix scoring*

For BERTScore differential matrix scoring, differential diagnoses were extracted from each model’s free-text clinical response using an open-source large language model (DeepSeek-LLM-7B-Chat). The model was prompted to return only differential diagnoses as a comma-separated list for parsing. Extracted differential diagnoses were compared with ground-truth differential lists using BERTScore with a ‘bert-base-multilingual-cased’ backbone. Precision was defined as the proportion of predicted diagnoses that matched a ground-truth diagnosis, recall as the proportion of ground-truth diagnoses recovered by the prediction, and F1 as the harmonic mean. For each model–dataset combination, precision, recall, and F1 were averaged across cases to obtain aggregate performance metrics.

```{text}
Given the following clinical response, extract ONLY the differential diagnoses.
Look for sections mentioning "differential diagnosis", "other possible diagnoses", etc.
Return ONLY a comma-separated list of differential diagnoses, wrapped in <DIFFERENTIALS> tags for easy parsing.
Format: <DIFFERENTIALS>Diagnosis 1, Diagnosis 2, Diagnosis 3</DIFFERENTIALS>

Clinical Response:
{response_text}

Differential Diagnoses:
```

```

*LLM-as-a-judge assessment*

Model-generated free-text outputs were post-processed using an LLM-assisted pipeline (Azure OpenAI deployment: gpt-4o; temperature=0) to (1) extract explicit disease diagnoses, (2) evaluate synonym-level agreement between model predictions and gold-standard diagnoses, and (3) perform severity triage against a predefined list of high-acuity dermatologic diagnoses.
Diagnosis extraction was performed using an LLM to identify explicit disease diagnoses from free-text outputs. Extracted diagnoses were parsed into structured lists using simple rule-based post-processing. Diagnosis matching (for the first 3 dermatologic diagnoses output in the response) assessed whether predicted diagnoses overlapped with reference diagnoses, allowing for synonymous or equivalent clinical entities. Binary match outcomes were derived from LLM judgments and encoded as indicator variables.
For severity assessment, predicted diagnoses were compared against a predefined list of severe dermatologic conditions. The LLM returned a similarity score and the closest matching severe diagnosis, or “NONE” if no match, which was parsed using deterministic rules.

"""
You are a board-certified dermatologist.

Input:
- Text that may include diagnoses, descriptions, morphologies, probabilities, or irrelevant content

Task:
- Extract ONLY explicit disease diagnoses
- Ignore image descriptions, anatomy, lesion terms (e.g. papule, pustule), probabilities, disclaimers, and duplicates
- If no valid disease diagnoses are present, return an empty list

Output:
- One diagnosis per line
- No numbering, explanations, or extra text

Text:
{input_text}
"""

"""
Input:
- A list of model-generated diagnoses
- A list of gold standard diagnoses
```

Task:
- For EACH model diagnosis, determine whether it represents the SAME clinical entity
  as ANY of the gold standard diagnoses (or an accepted synonym/variant)

Output:
- Respond with "Yes" or "No" on separate lines
- Preserve the original order
- No explanations or extra text
- Do NOT restate the diagnoses

Model diagnoses:
{diagnosis_lines}

Gold standard diagnoses:
{gold_lines}
"""

"""
Compare the predicted diagnosis(es) to the Severe Diagnosis List.
Return ONLY the following for the closest match:
1. A similarity score from 1 to 5 (1 = no match, 5 = identical)
2. The closest matching diagnosis from the Severe Diagnosis List — only the diagnosis name itself, no labels or prefixes. If there is no match, respond with "NONE".

Severe Diagnosis List:
- [list of severe diagnoses]

Predicted diagnosis(es):
{predicted_text}
""”

*Clinician evaluation (dermatologist-as-judge) analysis*

Clinician (dermatologist-as-judge) evaluation was completed using Label Studio. A subset of images and clinical prompts was selected randomly then screened iteratively to ensure no images contained visual PHI. That is, cases containing identifiable or potentially identifiable visual PHI, non-clinical imagery, or otherwise unsuitable content were excluded.

We assessed concordance between a subset of LLM-as-a-judge and clinician evaluation (dermatologist-as-judge) answers by assessing sensitivity, specificity and F1 scores. This was done for both top-3 accuracy and also severity assignment (value greater than 1 as severe) as binary labels.

For clinician (dermatologist-as-judge) evaluation, Responses from each MLLM architecture (with and without clinical context when relevant) were provided along with the original full prompt, the gold standard diagnosis/diagnoses, and the associated image. The physician was then asked to evaluate accuracy (whether any correct diagnosis is found in the response list) and severity (whether any diagnosis belongs to the list of severe diagnoses) corresponding to the same tasks requested for LLM-as-a-judge. Unlike LLM-as-a-judge, however, the physician was granted access to the full diagnosis list output by the MLLM rather than just the first three mentioned.

In addition, as part of this evaluation, the reasoning of each model response was also assessed. Model response visual-text reasoning was evaluated by physicians using predefined 5-point Likert scales across three dimensions: intent alignment, descriptive/medical knowledge factuality, and specificity to the reference diagnosis.

- Intent alignment assessed whether the response fulfilled the prompt requirements, including the presence, number, and validity of diagnoses. Higher scores reflected complete and accurate task fulfillment (without hallucination).
- Descriptive and medical knowledge factuality evaluated the accuracy of non-diagnostic content independent of diagnostic correctness, with scores reflecting the extent of factual inaccuracies.
- Specificity to the correct diagnosis assessed whether the response emphasized features (including but not limited to visual features) relevant to the reference diagnosis, ranging from irrelevant descriptions to comprehensive diagnostic specificity.

A rubric was also provided to each dermatologist to reduce variation between responses.

Unlike LLM-as-a-judge, clinician evaluations involved the relevant image used to generate the MLLM evaluation. While irrelevant to diagnosis accuracy assignment and severity assessment, image information was used when evaluating across the three reasoning domains to determine if the correct features of an image were being accurately described (factuality) and the relevant features were being identified (specificity).

*Inter-rater reliability analysis*

A subset of cases in each dataset was independently evaluated by three expert raters. Each unit of analysis was defined as a unique image and model response pair. Inter-rater agreement was quantified using the intraclass correlation coefficient (ICC), calculated as a two-way random-effects model assessing absolute agreement for raters. Binary outcomes (diagnosis correctness) were coded as 0 (incorrect) and 1 (correct) prior to analysis. Likert-scale outcomes were treated as continuous variables. Units with incomplete ratings were excluded using listwise deletion to ensure balanced rater contributions per unit. Ninety-five percent confidence intervals were calculated for each ICC estimate.

ICC values were interpreted using conventional thresholds: <0.40 poor, 0.40–0.59 fair, 0.60–0.74 good, 0.75–0.90 very good, and >0.90 excellent agreement.

*Statistics*

For Likert-scale means, we reported the mean and a 95% confidence interval (CI) computed as mean ± standard error, with 95% confidence intervals calculated using the t-distribution ($n-1$ degrees of freedom), using the sample SD and n within each stratum. To avoid unstable intervals in very small strata, CIs were suppressed when $n < 3$. Because scores were bounded (1–5), CI limits were clipped to [1, 5].

For proportions (such as for accuracy), 95% confidence intervals were computed using the Wilson score method for binomial outcomes ($k/n$), with bounds constrained to [0, 1]. Proportions and their corresponding Wilson confidence intervals were reported for each threshold k and group.